\documentclass{article}

\usepackage{microtype}
\usepackage{graphicx}
\usepackage{subfigure}
\usepackage{booktabs} % for professional tables

\usepackage{url}
\usepackage{graphicx}
\usepackage{amssymb,amsmath,amsthm,amsfonts,bm}
\usepackage{tabularx}
\usepackage{multirow}
\usepackage{diagbox}
\usepackage{wrapfig}
\usepackage{float}
\usepackage{tikz}
\usetikzlibrary{bayesnet}
\DeclareMathOperator*{\argmax}{arg\,max}
\DeclareMathOperator*{\argmin}{arg\,min}

\usepackage[accepted]{icml2018}

\icmltitlerunning{Inference Suboptimality in Variational Autoencoders}

\begin{document}

\twocolumn[
\icmltitle{Inference Suboptimality in Variational Autoencoders}

\icmlsetsymbol{equal}{*}

% \begin{icmlauthorlist}
% \icmlauthor{Aeiau Zzzz}{equal,to}
% \icmlauthor{Bauiu C.~Yyyy}{equal,to}
% \icmlauthor{Cieua Vvvvv}{goo}
% \end{icmlauthorlist}

\begin{icmlauthorlist}
\icmlauthor{Chris Cremer}{to}
\icmlauthor{Xuechen Li}{to}
\icmlauthor{David Duvenaud}{to}
\end{icmlauthorlist}

\icmlaffiliation{to}{Department of Computer Science, University of Toronto, Toronto, Canada}
% \icmlaffiliation{goo}{Googol ShallowMind, New London, Michigan, USA}
% \icmlaffiliation{ed}{School of Computation, University of Edenborrow, Edenborrow, United Kingdom}

\icmlcorrespondingauthor{Chris Cremer}{ccremer@cs.toronto.edu}

% \icmlcorrespondingauthor{Eee Pppp}{ep@eden.co.uk}


\vskip 0.3in
]



\printAffiliationsAndNotice{}  % leave blank if no need to mention equal contribution

\begin{abstract}
Amortized inference allows latent-variable models trained via variational learning to scale to large datasets.
The quality of approximate inference is determined by two factors: a) the capacity of the variational distribution to match the true posterior and b) the ability of the recognition network to produce good variational parameters for each datapoint. 
We examine approximate inference in variational autoencoders in terms of these factors. 
We find that divergence from the true posterior is often due to imperfect recognition networks, rather than the limited complexity of the approximating distribution.
We show that this is due partly to the generator learning to accommodate the choice of approximation.
Furthermore, we show that the parameters used to increase the expressiveness of the approximation play a role in generalizing inference rather than simply improving the complexity of the approximation.
\end{abstract}

\section{Introduction}
% Improving inference in variational autoencoders (VAEs)~\citep{vae,vae2} through the development of expressive approximate posteriors~\citep{normflow,iaf,hvm,householder,2017arXiv170602326T} improves the learned distribution over the data.
% Suboptimal inference leads to suboptimal models.
% %
% \vspace{-3mm}

% In this paper, we analyze inference suboptimality in VAEs: the mismatch between the true and approximate posterior. In other words, we are interested in understanding what factors cause the gap between the marginal log-likelihood and the evidence lower bound (ELBO). We refer to this as the inference gap.

\begin{figure}[h]
\begin{center}
\vspace{-1mm}
\includegraphics[width=.39\textwidth, clip, trim=0cm 0cm 0cm .2cm]{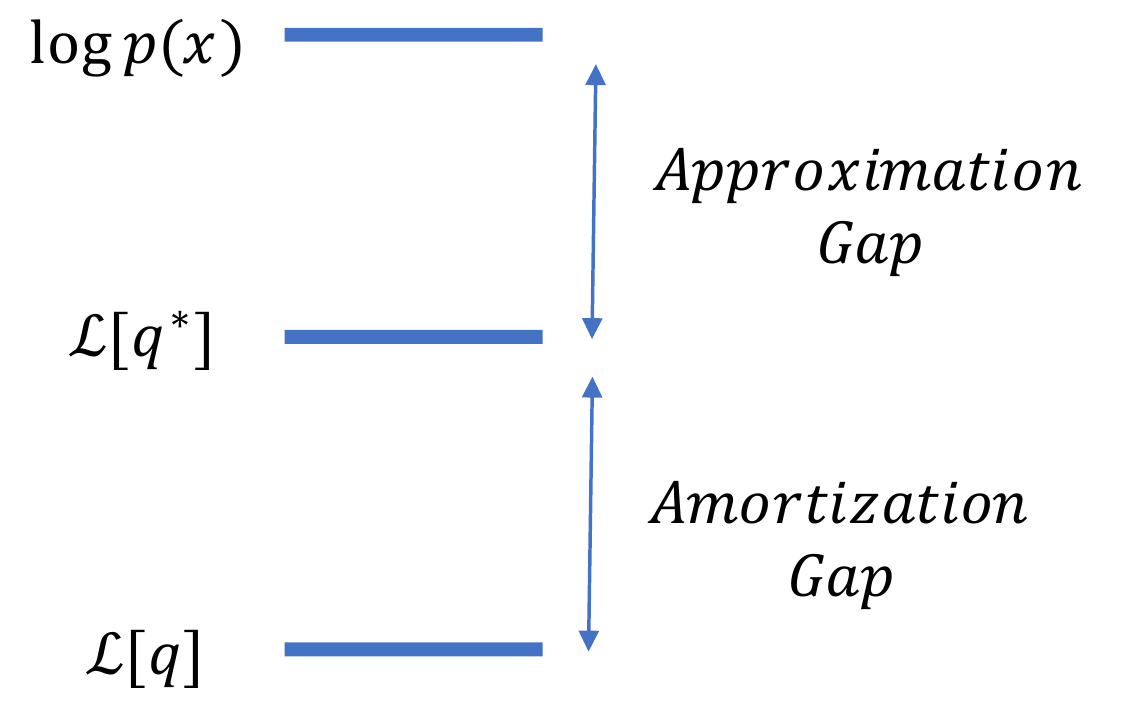}
\end{center}
% \caption{Gaps between optimal and approximate inference}
\caption{Gaps in Inference}
\label{gaps}
% \end{wrapfigure}
\end{figure}

% We break down the inference gap into two components: the \textit{approximation gap} and the \textit{amortization gap}.
% The approximation gap comes from the inability of the approximate distribution 
% %family 
% to exactly match the true posterior.
% The amortization gap refers to the difference caused by amortizing the variational parameters over the entire training set, instead of optimizing for each datapoint independently.
% We refer the reader to Table~\ref{summary_table} for detailed definitions and Figure~\ref{gaps} for a simple illustration of the gaps. In Figure~\ref{gaps}, $\mathcal{L}[q]$ refers to the ELBO using an amortized distribution $q$, whereas $q^*$ is the optimal within its variational family.
% 
In this paper, we analyze inference suboptimality: the mismatch between the true and approximate posterior.
More specifically, we are interested in understanding what factors cause the gap between the marginal log-likelihood and the evidence lower bound (ELBO) in variational autoencoders (VAEs, \citet{vae,vae2}). We refer to this as the \textit{inference gap}.
Moreover, we break down the inference gap into two components: the \textit{approximation gap} and the \textit{amortization gap}.
The approximation gap comes from the inability of the variational distribution family to %provide a candidate approximation to 
exactly match the true posterior.
The amortization gap refers to the difference caused by amortizing the variational parameters over the entire training set, instead of optimizing for each training example individually.
We refer the reader to Table~\ref{summary_table} for the definitions of the gaps and to Fig. \ref{gaps} for a simple illustration of the gaps. In Fig. \ref{gaps}, $\mathcal{L}[q]$ refers to the ELBO evaluated using an amortized distribution $q$, as is typical of VAE training. In contrast, $\mathcal{L}[q^*]$ is the ELBO evaluated using the optimal approximation within its variational family.

There has been significant work on improving variational inference in VAEs through the development of expressive approximate posteriors \citep{normflow,iaf,hvm,householder,2017arXiv170602326T}.
These works have shown that with more expressive approximate posteriors, the model learns a better distribution over the data. Our study aims to gain a better understanding of the relationship between expressive approximations and improved generative models.

Our experiments investigate how the choice of encoder, posterior approximation, decoder, and optimization affect the approximation and amortization gaps. We train VAE models in a number of settings on the MNIST \citep{mnist}, Fashion-MNIST \citep{fashion}, and CIFAR-10 \cite{krizhevsky2009learning} datasets.

% To understand what roles these gaps play in inference, we train VAE models in a number of settings.
% We compare the gaps under different approximate posterior families, different encoder sizes on the MNIST and Fashion-MNIST \citep{fashion} datasets.

Our contributions are:
a) we investigate inference suboptimality in terms of the approximation and amortization gaps, providing insight to guide future improvements in VAE inference, 
b) we quantitatively demonstrate that the learned generative model accommodates the choice of approximation,
% c) we show that highly parameterized approximate distributions reduce both the approximation and amortization gaps
and c) we demonstrate that parameterized functions that improve the expressiveness of the approximation play a significant role in reducing amortization error.

% \begin{wrapfigure}{ro}{0.39\textwidth}

\begin{table*}[t]
\centering
\begin{tabular}{clclc}
Term              &  & Definition                          &  &  VAE Formulation                                                       \\ \cline{1-1} \cline{3-3} \cline{5-5} 
Inference     &  & $\log p(x) - \mathcal{L}[q]$        &  & $\text{KL}\left( q(z|x) || p(z|x) \right)$                             \\
Approximation  &  & $\log p(x) - \mathcal{L}[q^*]$      &  & $\text{KL}\left( q^*(z|x) || p(z|x) \right)$                           \\
Amortization   &  & $\mathcal{L}[q^*] - \mathcal{L}[q]$ &  & $\text{KL}\left( q(z|x)||p(z|x) \right) - \text{KL}\left( q^*(z|x)||p(z|x) \right)$
\end{tabular}
\caption{Summary of Gap Terms. The middle column refers to the general case where our variational objective is a lower bound on the marginal log-likelihood. The right most column demonstrates the specific case in VAEs. $q^*(z|x)$ refers to the optimal approximation within a family $\mathcal{Q}$, i.e. $q^*(z|x) = \argmin_{q \in \mathcal{Q}} \text{KL}\left( q(z|x) || p(z|x) \right)$.} \label{summary_table}
\end{table*}

\section{Background}
\subsection{Inference in Variational Autoencoders}
Let $x$ be the observed variable, $z$ the latent variable, and $p(x, z)$ be their joint distribution. Given a dataset
$X=\{x_1, x_2, ..., x_N \}$,
we would like to maximize the marginal log-likelihood with respect to the model parameters $\theta$:
\begin{align}
    \log p_{\theta}(X) = \sum_{i=1}^N \log p_{\theta}(x_i) = \sum_{i=1}^N \log \int p_{\theta}(x_i, z_i) dz_i. \nonumber
\end{align}
In practice, the marginal log-likelihood is computationally intractable due to the integration over the latent variable $z$. 
% Instead, VAEs optimize the ELBO with respect to the model $p_{\theta}(x,z)$ parameterized by $\theta$ and the inference network $q_{\phi}(z|x)$ parameterized by $\phi$:
% We omit the parameterization subscripts for brevity. 
Instead, VAEs introduce an inference network $q_{\phi}(z|x)$ to approximate the true posterior $p(z|x)$ and optimize the ELBO with respect to model parameters $\theta$ and inference network parameters $\phi$  (parameterization subscripts omitted for brevity): 
\begin{align}
    \log p(x) &=  \mathbb{E}_{q(z|x)} \left[ \log \left(\frac{p(x,z)}{q(z|x)} \right)  \right] + \text{KL}\left( q(z|x)||p(z|x) \right)  \label{logmarginal}  \\
    &\geq  \mathbb{E}_{q(z|x)} \left[ \log \left(\frac{p(x,z)}{q(z|x)} \right)  \right]  = \mathcal{L}_{\text{VAE}}[q]. \label{vae_elbo}
\end{align}
From the above equation, we see that the ELBO is tight when $q(z|x) = p(z|x)$.
The choice of $q(z|x)$ is often a factorized Gaussian distribution for its simplicity and efficiency. 
% VAEs perform amortized inference by utilizing a recognition network (encoder), resulting in efficient approximate inference for large datasets. 
By utilizing the inference network (also referred to as encoder or recognition network), VAEs amortize inference over the entire dataset. 
% which  by utilizing a recognition network (encoder), resulting in efficient approximate inference for large datasets. 
Furthermore, the overall model is trained by stochastically optimizing the ELBO using the reparametrization trick \citep{vae}. 

% From the above equation, we see that the lower bound whenis tight if $q(z|x) = p(z|x)$.
% % The original VAE parameterizes the likelihood $p(x|z)$ and variational distribution $q(z|x)$ by deep neural nets. 

% \begin{figure}[!t]
%   \centering
%       \subfloat[Full dataset]
%         {
%         \begin{tabular}{lllll}
%         \multicolumn{1}{c}{Term} &  & \multicolumn{1}{c}{Definition} &  & \multicolumn{1}{c}{Compute}  \\ \cline{1-1}  \cline{3-3} \cline{5-5} 
%         Inference Gap            &  & $KL(q(z|x) || p(z|x))$         &  & AIS $-$ VAE[q]                 \\
%         Approximation Gap        &  & $KL(q^*(z|x) || p(z|x))$       &  & AIS $-$ VAE[q*]               \\
%         Amortization Gap         &  & $KL(q(z|x) || q^*(z|x))$       &  & VAE[q*]  $-$ VAE[q]
%         \end{tabular}
%         }
%       \subfloat[1000 datapoints]
%         {
%         \includegraphics[width=.1 \textwidth]{figs/smallN.pdf} \label{smallN}
%         }
%       \caption{Training curves for a FFG and a Flow inference model on MNIST.}
%     %   \label{ais_iw}
% \end{figure}

\subsection{Expressive Approximate Posteriors}

% In the original VAE, the inference distribution $q(z|x)$ can also be viewed as an approximation to the true posterior $p(z|x)$. This motivates the decomposition 

There are a number of strategies for increasing the expressiveness of approximate posteriors, going beyond the original factorized-Gaussian. We briefly summarize  normalizing flows and auxiliary variables.

\subsubsection{Normalizing Flows}
Normalizing flow \citep{normflow} is a change of variables procedure for constructing complex distributions by transforming probability densities through a series of invertible mappings. Specifically, if we transform a random variable $z_0$ with distribution $q_0(z)$, the resulting random variable $z_T=T(z_0)$ has a distribution: 
\begin{align}
    q_T(z_T) &= q_0(z_0)  \left | \mathrm{det} \frac{\partial z_T}{\partial z_0} \right |^{-1}. \label{change_of_var}
\end{align}
% A change of variable procedure such as normalizing flows \citep{normflow} is a tool for constructing complex distributions by transforming probability densities through a series of invertible mappings. More specifically, if we transform a random variable $z_0$ with distribution $q_0(z)$, the resulting random variable $z_T=T(z_0)$ has a distribution: 
% \begin{align}
%     q_T(z_T) &= q_0(z_0)  \left | \mathrm{det} \frac{\partial z_T}{\partial z_0} \right |^{-1}
% \end{align}
 By successively applying these transformations, we can build arbitrarily complex distributions. Stacking these transformations remains tractable due to the determinant being decomposable: 
 $\mathrm{det}(A B) = \mathrm{det}(A)\mathrm{det}(B)$. An important property of these transformations is that we can take expectations with respect to the transformed density $q_T(z_T)$ without explicitly knowing its formula due to the law of the unconscious statistician (LOTUS):
\begin{align}
    \mathbb{E}_{q_T}[h(z_T)] &= \mathbb{E}_{q_0}[h(f_T(f_{T-1}(...f_1(z_0))))].\label{lotus}
\end{align}
%This is known as the law of the unconscious statistician (LOTUS). 
% Using the change of variable and LOTUS, the lower bound can be written as:
% Using the change of variable formula
% % and the law of the unconscious statistician (LOTUS)
% and the LOTUS, the lower bound can be written as:
Using equations (\ref{change_of_var}) and (\ref{lotus}), the lower bound with the transformed approximation can be written as:
\begin{align}
    % \log p(x) & \geq 
    \mathbb{E}_{z_0 \sim q_0(z|x)} \left[\log\left(  \frac{p(x,z_T)}{q_0(z_0|x)\prod_{t=1}^{T} \left | \mathrm{det} \frac{\partial z_t}{\partial z_{t-1}} \right |^{-1}}  \right)  \right].\label{flow_elbo}
\end{align}
The main constraint on these transformations is that the determinant of their Jacobian needs to be easily computable.

\subsubsection{Auxiliary Variables} \label{aux_section}

Deep generative models can be extended with auxiliary variables which leave the generative model unchanged but make the variational distribution more expressive. Just as hierarchical Bayesian models induce dependencies between data, hierarchical variational models can induce dependencies between latent variables. 
% They can capture structure of correlated variables because they turn the posterior into a mixture of distributions: $q(z|x) = \int q(z|x,v)q(v|x)dz$.
% \begin{align}
%     q(z|x) &= \int q(z|x,v)q(v|x)dz
% \end{align}
  The addition of the auxiliary variable changes the lower bound to: 
\begin{align}
    % \log p(x) & \geq 
    & \mathbb{E}_{z,v \sim q(z,v|x)} \left[\log\left(  \frac{p(x,z)r(v|x,z)}{q(z, v|x)}  \right)  \right] \\
    &= \mathbb{E}_{q(z|x)} \left [\log\left(  \frac{p(x,z)}{q(z|x)}  \right) - \text{KL}\Bigl(q(v|z,x)\|r(v|x,z)\Bigr) \right]  \label{loose}%- \mathbb{E}_{ q(z|x)} \left [\text{KL}(q(v|z,x)\|r(v|x,z))  \right] \label{loose},
\end{align}
where $r(v|x,z)$ is called the reverse model. From Eqn. \ref{loose}, we see that this bound is looser than the regular ELBO, however the extra flexibility provided by the auxiliary variable can result in a higher lower bound. This idea has been employed in works such as auxiliary deep generative models (ADGM, \cite{adgm}),  hierarchical variational models (HVM, \cite{hvm}) and Hamiltonian variational inference (HVI, \cite{hvi}).

% % \subsection{Evaluation Metrics}
% \subsection{Marginal Log-Likelihood Estimation}

% % We are interested in estimating the marginal log likelihood of a model: $\log p(x) = \log \int p(x|z) p(z) dz$. 
% %a model assigns to an observation $x$. 
% % It is not easy to compute the marginal log likelihood of these models. We could do importance sampling using the prior but that has high variance. 

% % Here we describe two options for estimating the marginal log-likelihood of a model: IWAE and AIS.

% We use two bounds to estimate the marginal log-likelihood of a model: IWAE \citep{iwae} and AIS \citep{ais}. Here we describe the IWAE bound. See Section \ref{ais_section} in the Supplementary material for a description of AIS.

% % \subsubsection{IWAE Bound}

% The IWAE bound is a tighter lower bound than the VAE bound. More specifically, if we take multiple samples from the $q$ distribution, we can compute a tighter lower bound on the marginal log-likelihood:
% \begin{align}
%     \log p(x) & \geq \mathbb{E}_{z_{1}...z_{k} \sim q(z|x)} \left[\log\left(  \frac{1}{k}  \sum_{i=1}^k \frac{p(x,z_i)}{q(z_i|x)}  \right)  \right]\\
%     &= \mathcal{L}_{\text{IWAE}}[q]. \label{iwae_elbo} % %\label{iwae_elbo}  \eqname{(IWAE ELBO)}
% \end{align} 
% As the number of importance samples approaches infinity, the bound approaches the marginal log-likelihood. This importance weighted bound was introduced along with the Importance Weighted Autoencoder \citep{iwae}, thus we refer to it as the IWAE bound. It is often used as an evaluation metric for generative models \citep{iwae,iaf}. 

\section{Methods}

\subsection{Approximation and Amortization Gaps} \label{gaps_section}
The inference gap $\mathcal{G}$ is the difference between the marginal log-likelihood  $\log p(x)$ and a lower bound $\mathcal{L}[q]$.
Given the distribution in the family that maximizes the bound, $q^*(z|x) = \argmax_{q\in \mathcal{Q}} \mathcal{L}[q]$, the inference gap decomposes as the sum of approximation and amortization gaps:
\begin{align}
\mathcal{G} &= \log p(x) - \mathcal{L}[q] 
= \underbrace {\log p(x) - \mathcal{L}[q^*]}_{\text{Approximation}} + \underbrace{\mathcal{L}[q^*] - \mathcal{L}[q]}_{\text{Amortization}} .\nonumber
\end{align}
For VAEs, we can translate the gaps to KL divergences by rearranging Eqn. (\ref{logmarginal}):
\begin{align}
\mathcal{G}_{\text{VAE}} &= \underbrace{\text{KL}\bigl(q^*(z|x) || p(z|x)\bigr)}_{\text{Approximation}} \nonumber \\ &+  \underbrace{ \text{KL}\bigl( q(z|x)||p(z|x) \bigr) - \text{KL}\bigl(q^*(z|x) || p(z|x)\bigr) }_{\text{Amortization}}. 
\end{align}
% See Table \ref{summary_table} for a summary of these gaps. To obtain $q^*$, we optimize the parameters of the variational distribution for each datapoint independently.
% We emphasize that variational inference with a finite capacity inference model, even with perfect optimization, does not necessarily produce the optimal variational distribution due to performing amortization over the dataset. 
% In our experiments, we optimize the parameters of the variational distribution for each datapoint independently to reduce the gap caused by amortization and obtain $q^*$.

\subsection{Flexible Approximate Posteriors} \label{aux_nf}
Our experiments involve expressive approximations which use flow transformations and auxiliary variables. The flow transformation that we employ is of the same type as the transformations of Real NVP \citep{nvp}. We partition the latent variable $z$ into two, $z_1$ and $z_2$, then perform the following transformations:
\begin{align}
    z_1' &= z_1 \circ \sigma_1(z_2) + \mu_1(z_2) \label{trans1} \\
    z_2' &= z_2 \circ \sigma_2(z_1') + \mu_2(z_1')  \label{trans2}
\end{align}
where $\sigma_1, \sigma_2, \mu_1, \mu_2: \mathbb{R}^n \rightarrow \mathbb{R}^n$ are differentiable mappings parameterized by neural nets and $\circ$ takes the Hadamard or element-wise product. 
% The determinant of the combined transformation's Jacobian, $\left| \mathrm{det} \left(\frac{\partial z_tv_t}{\partial z_{t-1}v_{t-1}} \right) \right |$, can be easily evaluated. 
We partition the latent variable by simply indexing the elements of the first half and the second half. 
The determinant of the combined transformation's Jacobian, $\left| \mathrm{det} \left(\frac{\partial z'}{\partial z} \right) \right |$, can be easily evaluated. See section \ref{flow_det} of the Supplementary material for a derivation. The lower bound of this approximation is the same as Eqn. (\ref{flow_elbo}). We refer to this approximation as $q_{Flow}$.

We also experiment with an approximation that combines flow transformations and auxiliary variables. 
% We use two expressive distributions: $q_{Flow}$ and $q_{AF}$.
% the bound is teh smae as (6) for Flow. 
% Our experiments compare two families of approximate posteriors: the fully-factorized Gaussian (FFG) and a flexible flow (Flow). Our choice of flow is a combination of the Real NVP \citep{nvp} and auxiliary variables \citep{hvm, adgm}. 
% Our model also resembles leap-frog dynamics applied in Hamiltonian Monte Carlo (HMC, \citet{hmc}).
Let $z\in \mathbb{R}^n$ be the variable of interest and $v\in \mathbb{R}^n$ the auxiliary variable. 
The flow is the same as equations (\ref{trans1}) and (\ref{trans2}), where $z_1$ is replaced with $z$ and $z_2$ with $v$. We refer to this approximate distribution as $q_{AF}$, where AF stands for auxiliary flow. 
% Each flow step involves:
% \begin{align}
%     v' &= v \circ \sigma_1(z) + \mu_1(z) \label{trans1} \\
%     z' &= z \circ \sigma_2(v') + \mu_2(v')  \label{trans2}
% \end{align}
% Thus, we can jointly train the generative and flow-based inference model
We train this model by optimizing the following bound:
\begin{align}
    % \log p(x) &\geq 
    &\mathbb{E}_{q_0(z,v|x)} \left[\log\left(  \frac{p(x,z_T)r(v_T|x,z_T)}{q_T(z_T, v_T|x) \left 
    | \mathrm{det} \left(\frac{\partial z_tv_t}{\partial z_{t-1}v_{t-1}}\right) \right |^{-1}}  \right)  \right] \nonumber \\
    &= \mathcal{L}[q_{AF}]. 
\end{align}
Note that this lower bound is looser as explained in Section \ref{aux_section}.
% Additionally, multiple such type of transformations can be stacked to improve  expressiveness.
We refer readers to Section \ref{exps_details} in the Supplementary material for specific details of the flow configuration adopted in the experiments. 
% where each of the distributions are parameterized by neural nets. See the appendix for a description of the model architectures and hyperparameters. 

% The flow of Fig. \ref{toy} is a demonstration of the distribution approximation flexibility of this type of distribution.

\subsection{Marginal Log-Likelihood Estimation and Evidence Lower Bounds} \label{eval_bounds}

In this section, we describe the estimates we use to compute the bounds of the inference gaps: $\log p(x)$, $\mathcal{L}[q^*]$, and $\mathcal{L}[q]$. 
We use two bounds to estimate the marginal log-likelihood, $\log p(x)$: IWAE \citep{iwae} and AIS \citep{ais}.

% \subsection{Evaluation Metrics}
% \subsection{Marginal Log-Likelihood Estimation}

% We are interested in estimating the marginal log likelihood of a model: $\log p(x) = \log \int p(x|z) p(z) dz$. 
%a model assigns to an observation $x$. 
% It is not easy to compute the marginal log likelihood of these models. We could do importance sampling using the prior but that has high variance. 

% Here we describe two options for estimating the marginal log-likelihood of a model: IWAE and AIS.

% To estimate the marginal log-likelihood, $\log \hat{p}(x)$, we take the maximum of our tightest lower bounds, specifically the maximum between the IWAE and AIS bounds. 

% \subsubsection{IWAE Bound}

% The IWAE bound is a tighter lower bound than the VAE bound. More specifically, if we
The IWAE bound takes multiple importance weighted samples from the variational $q$ distribution resulting in a tighter lower bound than the VAE bound. 
% we can compute a tighter lower bound on the marginal log-likelihood:
The IWAE bound is computed as:
\begin{align}
    \log p(x) & \geq \mathbb{E}_{z_{1}...z_{k} \sim q(z|x)} \left[\log\left(  \frac{1}{k}  \sum_{i=1}^k \frac{p(x,z_i)}{q(z_i|x)}  \right)  \right]\\
    &= \mathcal{L}_{\text{IWAE}}[q]. \label{iwae_elbo} \nonumber  % %\label{iwae_elbo}  \eqname{(IWAE ELBO)}
\end{align} 
As the number of importance samples approaches infinity, the bound approaches the marginal log-likelihood. 
% This importance weighted bound was introduced along with the Importance Weighted Autoencoder \citep{iwae}, thus we refer to it as the IWAE bound. 
It is often used as an evaluation metric for generative models \citep{iwae,iaf}. 
% However it is encoder-dependent
AIS is potentially an even tighter lower bound. AIS weights samples from distributions which are sequentially annealed from an initial proposal distribution to the true posterior. See Section \ref{ais_section} in the Supplementary material for further details regarding AIS. 
To compute the AIS bound, 
%When computing the AIS bound, 
%$\mathcal{L}_{\text{AIS}}$, we use Hamiltonian Monte Carlo (HMC, \citet{hmc}) as our transition operator $\mathcal{T}_t(z'|z)$. 
we use 100 chains, each with 10000 intermediate distributions, where each transition consists of one HMC trajectory with 10 leapfrog steps. The initial distribution for AIS is the prior, so that it is encoder-independent.

We estimate the marginal log-likelihood by independently computing our tightest lower bounds then take the maximum of the two: $$\log \hat{p}(x) = \max(\mathcal{L}_{\text{AIS}}, \mathcal{L}_{\text{IWAE}}).$$
% For our experiments, we test two different variational distributions: the fully-factorized Gaussian $q_{FFG}$ and the flexible approximation $q_{Flow}$ as described in section \ref{aux_nf}.
% $\mathcal{L}_{\text{IWAE}}[q]$ is the bound defined in Eqn. \ref{iwae_elbo}. 
The $\mathcal{L}[q^*]$ and $\mathcal{L}[q]$ bounds are the standard ELBOs, $\mathcal{L}_{\text{VAE}}$, from Eqn. (\ref{vae_elbo}), computed with either the amortized $q$ or the optimal $q^*$ (see below).
When computing $\mathcal{L}_{\text{VAE}}$ and $\mathcal{L}_{\text{IWAE}}$, we use 5000 samples.
% We use three bounds: VAE, IWAE, and AIS.
% To compute the AIS bound,

\begin{figure*}[th]
  \centering
    %   \fbox{\includegraphics[width=1. \textwidth, clip, trim=1.5cm 1.7cm 1.2cm 1.8cm]{figs/rearranged_real_blue.pdf}}
    \qquad \qquad \qquad \qquad A \quad \qquad \qquad \qquad B \quad \qquad \qquad \qquad C \quad \qquad \qquad \qquad D
    \includegraphics[width=.84 \textwidth, clip, trim=1.cm 1.cm 1.cm .5cm]{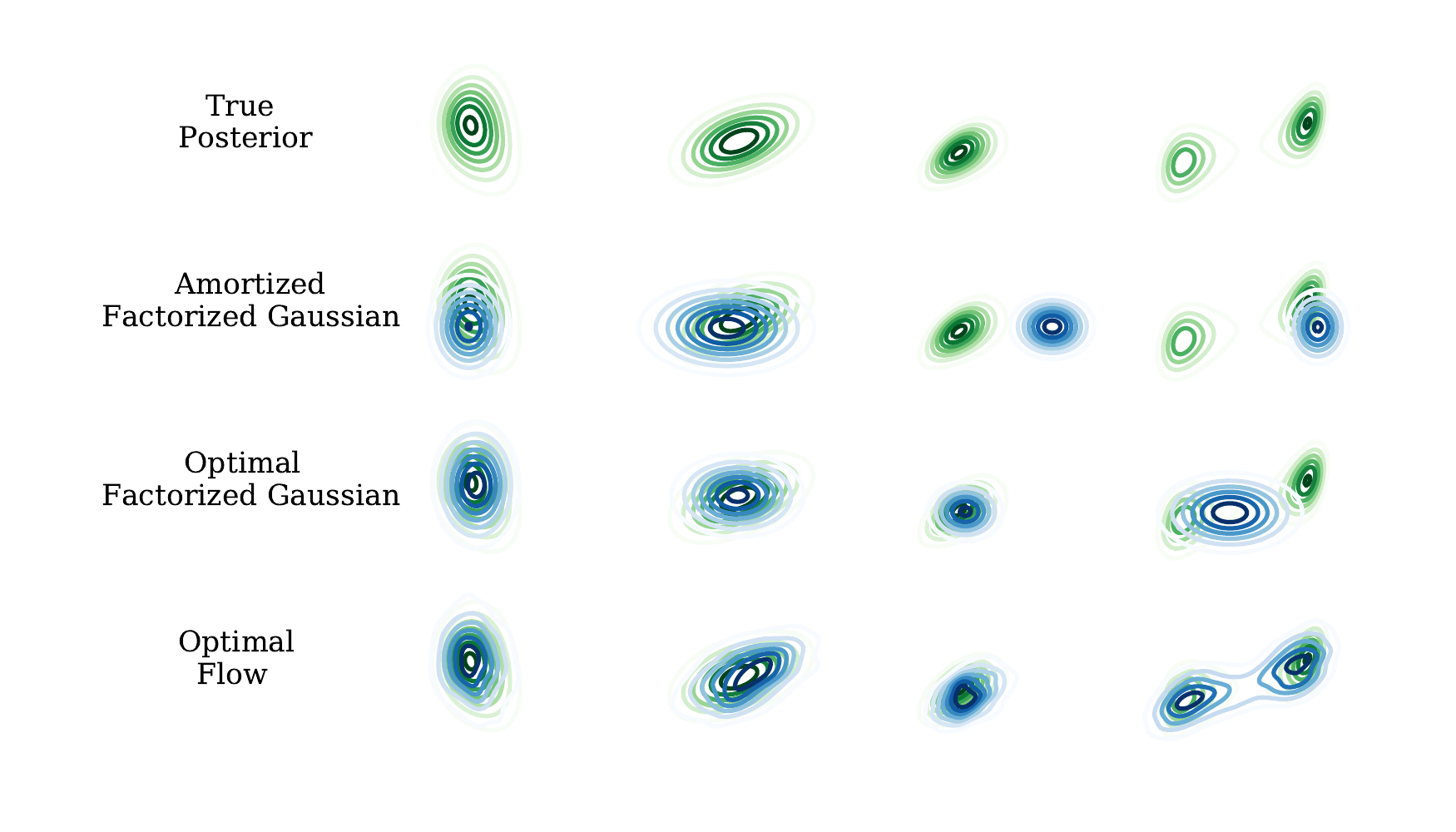}
  \caption{True Posterior and Approximate Distributions of a VAE with 2D latent space. The columns represent four different datapoints. The green distributions are the true posterior distributions, highlighting the mismatch with the blue approximations. Amortized: Variational parameters learned over the entire dataset. Optimal: Variational parameters optimized for each individual datapoint. Flow: Using a flexible approximate distribution.}
  \label{true_post}
\end{figure*}

% See Section \ref{local_opt_description} for details of the local optimization and stopping criteria.
\subsection{Local Optimization of the Approximate Distribution} \label{local_opt_description}

To compute $\mathcal{L}_{\text{VAE}}[q^*]$, we optimize the parameters of the variational distribution for every datapoint. 
For the local optimization of $q_{FFG}$, we initialize the mean and variance as the prior, i.e. $\mathcal{N}(0,I)$. We optimize the mean and variance using the Adam optimizer with a learning rate of $10^{-3}$.  % per optimization step. 
% to reduce variance. 
To determine convergence, after every 100 optimization steps, we compute the average of the  previous 100 ELBO values and compare it to the best achieved average. If it does not improve for 10 consecutive iterations then the optimization is terminated. For $q_{Flow}$ and $q_{AF}$, the same process is used to optimize all of its parameters. All neural nets for the flow were initialized with a variant of the Xavier initilization \citep{glorot2010understanding}. We use 100 Monte Carlo samples to compute the ELBO to reduce variance.

\subsection{Validation of Bounds}
The soundness of our empirical analysis depends on the reliability of the marginal log-likelihood estimator. For general importance sampling based estimators, the sample variance of the normalized importance weights can serve as an indicator of accuracy \citep{geweke1989bayesian, ais}. This quantitative measure, however, can also be unreliable, e.g. when the proposal misses an important mode of the target distribution \citep{ais}.

In this work, we follow \cite{ais_eval} to empirically validate our AIS estimates with Bidirectional Monte Carlo (BDMC, \citet{grosse2015sandwiching, grosse2016measuring}). In addition to a lower bound provided by AIS, BDMC runs AIS chains backward from exact posterior samples to obtain an upper bound on the marginal log-likelihood. 
% The idea is that we should be able to trust our AIS estimates when the upper and lower bounds roughly match (e.g. within 1 nat, \citep{grosse2015sandwiching}). 
It should be noted that BDMC relies on the assumption that the distribution of the simulated data from the model roughly matches that of the real data. This is due to the backward chain initializes from exact posterior samples \citep{grosse2015sandwiching}. 
% In general, this assumption does not hold, and we empirically observed considerable mismatch for the entropy in some cases.

For the MNIST and Fashion datasets, BDMC gives a gap within 0.1 nat for a linear schedule AIS with $10^4$ intermediate distributions and 100 importance samples on $10^3$ simulated datapoints. 
For 3-BIT CIFAR, the same AIS setting gives a gap within 1 nat with the sigmoidial annealing schedule \cite{grosse2015sandwiching} on 100 simulated datapoints. Loosely speaking, this should give us confidence in how well our AIS lower bounds reflect the marginal log-likelihood computed on the real data.

\begin{table*}[t]
\centering
\begin{tabular}{cccclccclccc}
\multicolumn{1}{c}{}                   & \multicolumn{3}{c}{MNIST} & \multicolumn{1}{c}{} & \multicolumn{3}{c}{Fashion-MNIST} & \multicolumn{1}{c}{} & \multicolumn{3}{c}{3-BIT CIFAR}\\ 
 \cline{2-4} \cline{6-8} \cline{10-12} 
   & $q_{FFG}$ &  & $q_{AF}$ & & $q_{FFG}$    &     & $q_{AF}$ & & $q_{FFG}$    &     & $q_{AF}$ \\
\cline{2-2} \cline{4-4} \cline{6-6} \cline{8-8} \cline{10-10} \cline{12-12}
$\log \hat{p}(x)$                      & -89.80    &  & -88.94  &   & -97.47 &     & -97.41 &   & -816.9 &     & -820.56    \\
$\mathcal{L}_{\text{VAE}}[q^*_{AF}]$ & -90.80    &  & -90.38  &   & -98.92 &     & -99.10 &   & -820.19 &     & -822.16    \\
$\mathcal{L}_{\text{VAE}}[q^*_{FFG}]$  & -91.23    &  & -113.54 &   & -100.53&     & -132.46&   & -831.65 &     & -861.62    \\
$\mathcal{L}_{\text{VAE}}[q]$          & -92.57    &  & -91.79  &   & -104.75&     & -103.76&   & -869.12 &     & -864.28    \\ \hline
Approximation                          & 1.43      &  & 1.44    &   & 3.06   &     & 1.69   &   & 14.75      &     & 1.60    \\
Amortization                           & 1.34      &  & 1.41    &   & 4.22   &     & 4.66   &   & 37.47     &     & 42.12    \\
Inference                              & 2.77      &  & 2.85    &   & 7.28   &     & 6.35   &   & 52.22     &     & 43.72  
\end{tabular}
\caption{Inference Gaps. The columns $q_{FFG}$ and $q_{AF}$ refer to the variational distribution used for training the model. These lower bounds are computed on the training set and are in units of nats.
}
\label{opt_table}
\end{table*}

\section{Related Work}
% Model evaluation with AIS appears early on in the setting of deep belief networks \citep{salakhutdinov2008quantitative}. \citet{ais_eval} first propose using AIS to evaluate decoder-based models, such as VAEs, GANs, and GMMNs. Their work also validates the accuracy of the approach with Bidirectional Monte Carlo (BDMC, \citet{grosse2015sandwiching}). In particular, they demonstrate the advantage of using AIS over the IWAE bound for evaluation when the inference network overfits to the training data.
Much of the earlier work on variational inference focused on optimizing the variational parameters locally for each datapoint, e.g. the original Stochastic Variational Inference scheme (SVI, \citet{hoffman2013stochastic}). To scale inference to large datasets, most related works utilize inference networks to amortize the cost of inference over the entire dataset.
Our work analyses the error that these inference networks introduce. 
% specifies the variational parameters to be optimized locally 
% in the inner loop.
% \citet{salakhutdinov2010efficient} perform such local optimization when learning deep Boltzmann machines. More recent work has applied this idea to improve approximate inference in directed Belief networks \citep{hjelm2015iterative}. 

% This suggests that inference potentially could be improved with better amortization, e.g. by further optimizing the local variational parameters during training \citep{introspective}. Moreover, we could further reduce the amortization error with flows by tailoring the flow that transforms the distribution $q(z|x)$ to the specific datapoint $x$ at hand. Inverse autoregressive flows (IAF, \citet{iaf}) leverage this information by making the flow model take in activations of some hidden layer of the encoder. 

% Pertaining to VAEs, the ladder variational autoencoder (LVAE, \citet{sonderby2016ladder}) learns approximate inference by modifying the generative distribution when there is a hierarchy of latent variables. This also suggests that amortized inference could induce an error which cannot be ignored.

Most relevant to our work is the recent work of \citet{2017arXiv171006085K}, which explicitly remarks on two sources of error in variational learning with inference networks, and proposes to optimize approximate inference locally from an initialization output by the inference network. 
They show improved training on high-dimensional, sparse data with the hybrid method, claiming that local optimization reduces the negative effects of random initialization in the inference network early on in training. Thus their work focuses on reducing the amortization gap early on in training. 
% and does not analyze the error arising from the use of limited approximating distributions.
Similar to this idea, \citet{hoffman2017learning} proposes to perform approximate inference during model training with MCMC at an initialization given by a variational distribution.
Our work provides a means of explaining these improvements in terms of the sources of inference suboptimality that they reduce.
% Markov Chain Monte Carlo (MCMC, \citet{brooks2011handbook}) at an initialization given by a variational distribution.

% Even though it is clear that failed inference would lead to a failed generative model, little quantitative assessment has been done showing the effect of the approximate posterior on the true posterior. \citet{iwae} visually demonstrate that when trained with an importance-weighted approximate posterior, the resulting true posterior is more complex than those trained with fully-factorized Gaussian approximations. We extend this observation quantitatively to the setting of flow-based approximate inference.

% Citation from Area chair: \cite{turner}

% Citation: L2HMC + other HMC related

\section{Experimental Results}

% Our choice of flexible inference is a combination of normalizing flows \citep{nvp} and auxiliary variables \citep{hvm}. Description:. maybe have methods section.

% \subsection{Inference Visualization}
\subsection{Intuition through Visualization}

To begin, we would like to gain an intuitive visualization of the gaps presented in Section \ref{gaps_section}.
% gain some insight into the properties of inference in VAEs by visualizing different distributions in the latent space. 
To this end, we trained a VAE with a two-dimensional latent space on MNIST and in Fig. \ref{true_post} we show contour plots of various distributions in the latent space. The first row contains contour plots of the true posteriors $p(z|x)$ for four different training datapoints (columns). We have selected these four examples to highlight different inference phenomena. The amortized fully-factorized Gaussian (FFG) row refers to the output of the recognition net, in this case, a FFG approximation. Optimal FFG is the FFG that best fits the posterior of the datapoint. Optimal Flow is the optimal fit of a flexible distribution to the same posterior, where the flexible distribution we use is described in Section \ref{aux_nf}.

% Fig. \ref{toy} is a demonstration of the distribution approximation flexibility of flows. We see that it can approximate arbitrarily complex distributions. 

% To gain insight into the properties of inference in VAEs, we will visualize inference by training a model on MNIST with a 2 dimensional latent space. Fig. \ref{real} are visualizations of distributions in the latent space. 'True' refers to the true posterior distribution $p(z|x)$. 'Amortized FFG' is the approximate posterior $q(z|x)$ used during training. 'Optimal Flow' is the flexible distribution optimized for each individual datapoint.

 Posterior A is an example of a distribution where a FFG can fit relatively well. Posterior B is an example of a posterior with dependence between dimensions, demonstrating the limitation of having a factorized approximation.
 %thus the FFG is limited by its factorization. 
 Posterior C highlights a shortcoming of performing amortization with a limited-capacity recognition network, where the amortized FFG shares little support with the true posterior. Posterior D is a bi-modal distribution which demonstrates the ability of the flexible approximation to fit to complex distributions, in contrast to the simple FFG approximation.
%  the limitations of the FFG approximation are more pronounced.
These observations raise the following question: in more typical VAEs, is the amortization of inference the leading cause of the distribution mismatch, or is it the limited expressiveness of the approximation? %This question will be explored in the following sections.

\begin{table*}[ht]
\centering
\begin{tabular}{cccclcccccccccccc}
\multicolumn{1}{l}{}                   & \multicolumn{3}{c}{MNIST} & \multicolumn{1}{c}{} & \multicolumn{3}{c}{Fashion-MNIST} &  &  & \multicolumn{3}{c}{MNIST} &  & \multicolumn{3}{c}{Fashion-MNIST} \\ \cline{2-4} \cline{6-8} \cline{11-13} \cline{15-17} 
                                       & $q_{FFG}$ &  & $q_{AF}$ &                      & $q_{FFG}$    &     & $q_{AF}$   &  &  & $q_{FFG}$ &  & $q_{AF}$ &  & $q_{FFG}$  &   & $q_{AF}$ \\ \cline{2-2} \cline{4-4} \cline{6-6} \cline{8-8} \cline{11-11} \cline{13-13} \cline{15-15} \cline{17-17} 
$\log \hat{p}(x)$                      & -89.61    &  & -88.99     &                      & -95.99       &     & -96.18       &  &  & -89.82    &  & -89.52     &  & -102.56    &   & -102.88    \\
$\mathcal{L}_{\text{VAE}}[q^*_{AF}]$ & -90.65    &  & -90.44     &                      & -97.40       &     & -97.91       &  &  & -90.96    &  & -90.45     &  & -103.73    &   & -104.02    \\
$\mathcal{L}_{\text{VAE}}[q^*_{FFG}]$  & -91.07    &  & -108.71    &                      & -99.64       &     & -129.70      &  &  & -90.84    &  & -92.25     &  & -103.85    &   & -105.80    \\
$\mathcal{L}_{\text{VAE}}[q]$          & -92.18    &  & -91.19     &                      & -102.73      &     & -101.67      &  &  & -92.33    &  & -91.75     &  & -106.90    &   & -107.01    \\ \cline{1-8} \cline{11-17} 
Approximation                          & 1.46      &  & 1.45       &                      & 3.65         &     & 1.73         &  &  & 1.02      &  & 0.93       &  & 1.29       &   & 1.14       \\
Amortization                           & 1.11      &  & 0.75       &                      & 3.09         &     & 3.76         &  &  & 1.49      &  & 1.30       &  & 3.05       &   & 2.29       \\
Inference                              & 2.56      &  & 2.20       &                      & 6.74         &     & 5.49         &  &  & 2.51      &  & 2.23       &  & 4.34       &   & 4.13      
\end{tabular}
\caption{Left: Larger Encoder. Right: Models trained without entropy annealing.
The columns $q_{FFG}$ and $q_{AF}$ refer to the variational distribution used for training the model. The lower bounds are computed on the training set and are in units of nats.}
% All numbers are in nats.}
\label{larger_encoder}
\end{table*}

\subsection{Amortization vs Approximation Gap} \label{amort_approx}

In this section, we compare how much the approximation and amortization gaps each contribute to the total inference gap. Table \ref{opt_table} are results of inference on the training set of MNIST, Fashion-MNIST and 3-BIT CIFAR (a binarized version of CIFAR-10, see Section \ref{3bitcifar} for details). For each dataset, we trained models with two different approximate posterior distributions: a fully-factorized Gaussian, $q_{FFG}$, and the flexible distribution, $q_{AF}$.
Due to the computational cost of optimizing the local parameters for each datapoint, our evaluation is performed on a subset of 1000 datapoints for MNIST and Fashion-MNIST and a subset of 100 datapoints for 3-BIT CIFAR.

For MNIST, we see that the amortization and approximation gaps each account for nearly half of the inference gap. On the more difficult Fashion-MNIST dataset, the amortization gap is larger than the approximation gap.
% the amortization gap is slightly larger than the approximation gap on the FFG-trained model. For the flow model, although the inference gap is slightly smaller than the FFG model, a larger portion of its inference gap comes from the approximation gap. 
For CIFAR, we see that the amortization gap is much more significant compared to the approximation gap. Thus, for the three datasets and model architectures that we consider, the amortization gap is likely to be the more prominent cause of inference suboptimality, especially when the dataset becomes more challenging to model. This indicates that improvements in inference will likely be a result of reducing amortization error, rather than approximation errors.

% Table \ref{fashion} are the results of the same experiment but on the Fashion MNIST dataset \citep{fashion}. 
% In this case, the amortization gap for both models is dominated by the amortization gap. This indicates that an encoder with larger capacity is required for improved inference, rather than increasing the expressiveness of the approximate distribution.

With these results in mind, would simply increasing the capacity of the encoder improve the amortization gap? We examined this by training the MNIST and Fashion-MNIST models from above but with larger encoders. See Section \ref{exps_details} for implementation details. Table \ref{larger_encoder} (left) are the results of this experiment. Comparing to Table \ref{opt_table}, we see that, for both datasets and both variational distributions, using a larger encoder results in the inference gap decreasing and the decrease is mainly due to a reduction in the amortization gap.

\subsection{Influence of Flows on the Amortization Gap} \label{flows_on_amort}

The common reasoning for increasing the expressiveness of the approximate posterior is to minimize the difference between the true and approximate distributions, i.e. reduce the approximation gap. However, given that the expressive approximation is often accompanied by many additional parameters, we would like to know how much influence it has on the amortization error.

To investigate this, we trained a VAE on MNIST, discarded the encoder, then retrained encoders with different approximate distributions on the fixed decoder. We fixed the decoder so that the true posterior is constant for all the retrained encoders. The initial encoder was a two-layer MLP with a factorized Gaussian distribution. In order to emphasize a large amortization gap, the retrained encoders had no hidden layers (ie. just linear transformations). For the retraiend encoders, we tested three approximate distributions: fully factorized Gaussian ($q_{FFG}$), auxiliary flow ($q_{AV}$), and Flow ($q_{Flow}$). See Section \ref{aux_nf} for the details of these distributions.

\begin{table}[ht]
\centering
\begin{tabular}{clccl}
Variational Family              &  & $q_{FFG}$ & $q_{AF}$ & $q_{Flow}$ \\ \cline{1-1} \cline{3-5} 
$\log \hat{p}(x)$               &  & -84.70    & -84.70   & -84.70     \\
$\mathcal{L}_{\text{VAE}}[q^*]$ &  & -86.61    & -85.48   & -85.13     \\
$\mathcal{L}_{\text{VAE}}[q]$   &  & -129.83   & -98.58   & -97.99     \\ \cline{1-1} \cline{3-5} 
Approximation                   &  & 1.91      & 0.78     & 0.43       \\
Amortization                    &  & 43.22     & 13.10    & 12.86      \\
Inference                       &  & 45.13     & 13.88    & 13.29     
\end{tabular}
\caption{Influence of expressive approximations on the amortization gap. The parameters used to increase the flexibility of the approximate distribution also reduce the amortization gap. }
% See Section \ref{flows_on_amort} for details of the experiment.}
\label{flow_affect}
\end{table}

The inference gaps of the retrained encoders on the training set are shown in Table \ref{flow_affect}. As expected, we observe that the small encoder with $q_{FFG}$ has a very large amortization gap. However, when we use $q_{AF}$ or $q_{Flow}$ as the approximate distribution, we see the approximation gap decrease, but more importantly, there is a significant decrease in the amortization gap. This indicates that the parameters used for increasing the complexity of the approximation also play a large role in diminishing the amortization error.

These results are expected given that the parameterization of the Flow distribution can be interpreted as an instance of the RevNet \citep{gomez2017reversible} which has demonstrated that Real-NVP transformations \citep{nvp} can model complex functions similar to typical MLPs. Thus the flow transformations we employ should also be expected to increase the expressiveness while also increasing the capacity of the encoder. The implication of this observation is that models which improve the flexibility of their variational approximation, and attribute their improved results to the increased expressiveness, may have actually been due to the reduction in amortization error. %rather than approximation.

% THE POINT IS THAT THE IMPROVED RESULTS of .. may have been attributed/justify to should have been attributed to the increased capacity 

\begin{figure*}[th]
  \centering
    %   \fbox{\includegraphics[width=1. \textwidth, clip, trim=1.5cm 1.7cm 1.2cm 1.8cm]{figs/rearranged_real_blue.pdf}}
    \includegraphics[width=1. \textwidth, clip, trim=1.cm 1.cm 1.cm 0.cm]{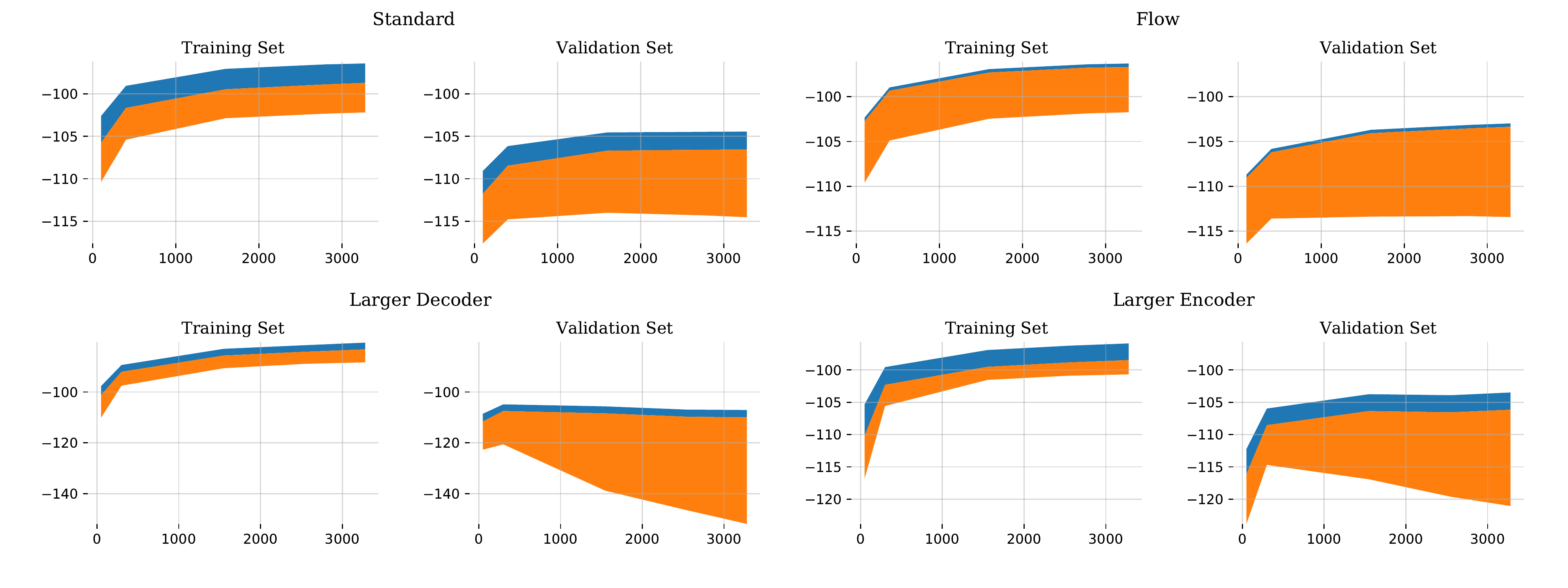} %8plotsgaps
  \caption{Inference gaps over epochs trained on binarized Fashion-MNIST. Blue is the approximation gap. Orange is the amortization gap. Standard is a VAE with FFG approximation. Flow is a VAE with a Flow approximation.}
  \label{gaps_over_epochs}
\end{figure*}

\subsection{Influence of Approximate Posterior on True Posterior} \label{approx_on_true}

%We have seen that increasing the expressiveness of the approximation improves the marginal likelihood of the trained model, but to what extent does it alter the true posterior?
To what extent does the posterior approximation affect the learned model?
\citet{turner} studied the biases in parameter learning induced by the variational approximation when learning via variational Expectation-Maximization.
% that the variational approximation affects the parameter estimates returned by variational E
Similarly, we ask whether a factorized Gaussian approximation causes the true posterior to be more like a factorized Gaussian?
%or is the true posterior mostly fixed?
\citet{iwae} visually demonstrate that when trained with an importance-weighted approximate posterior, the resulting true posterior is more complex than those trained with factorized Gaussian approximations. Just as it is hard to evaluate a generative model by visually inspecting samples, it is hard to judge how ``Gaussian" the true posterior is by visual inspection. We can quantitatively determine how close the posterior is to a fully-factorized Gaussian (FFG) by comparing 
% the Optimal FFG bound, $\mathcal{L}_{\text{VAE}}[q^*_{FFG}]$, and the Optimal Flow bound, $\mathcal{L}_{\text{VAE}}[q^*_{Flow}]$. 
the marginal log-likelihood estimate $\log \hat{p}(x)$ and the Optimal FFG bound $\mathcal{L}_{\text{VAE}}[q^*_{FFG}]$. This is equivalent to estimating the KL divergence between the optimal Gaussian and the true posterior, $\text{KL}\left( q^*(z|x) || p(z|x) \right)$. 

In Table \ref{opt_table} on MNIST,
% $\mathcal{L}_{\text{VAE}}[q^*_{AF}]$ improves upon the $\mathcal{L}_{\text{VAE}}[q^*_{FFG}]$ for the FFG trained model by 0.4 nats
% the $\text{KL}\left( q_{FFG}^*(z|x) || p(z|x) \right)$ and $\text{KL}\left( q_{AF}^*(z|x) || p(z|x) \right)$ 
for the FFG trained model,  $\text{KL}\left( q^*(z|x) || p(z|x) \right)$ is nearly the same for both $q_{FFG}^*$ and $q_{AF}^*$. In contrast, on the model trained with $q_{AF}$, $\text{KL}\left( q^*(z|x) || p(z|x) \right)$ is much larger for $q_{FFG}^*$ than $q_{AF}^*$.
% the difference increases to 12.5 nats. 
This suggests that the true posterior of a FFG-trained model is closer to FFG than the true posterior of the Flow-trained model. The same observation can be made on the Fashion-MNIST dataset. This implies that the decoder can learn to have a true posterior that fits better to the approximation.
% Although the generative model can learn to have a posterior that fits to the approximation, it seems that not having this constraint, ie. using a flexible approximate, results in better generative models.
% The same observation can be seen on the Fashion MNIST dataset in Table \ref{fashion}. 

% This indicates that the true posterior of the Flow-trained model with warmup is not a very good fit for a FFG distribution.

% The implication on these results is tha
% t  gaps: is that the approximation gap wont be too bad 

%We can use these observations to help justify our approximation and amortization gap results of Section \ref{amort_approx}.
These observations justify our results of Section \ref{amort_approx}.
which showed that the amortization error is often the main cause of inference suboptimality. One reason for this is that the generator accommodates the choice of approximation, thus reducing the approximation error. 

\begin{table}[ht]
\centering
\begin{tabular}{clccc}
Generator Hidden Layers               &  & 0       & 2      & 4      \\ \cline{1-1} \cline{3-5} 
$\log \hat{p}(x)$                     &  & -100.52 & -84.78 & -82.19 \\
$\mathcal{L}_{\text{VAE}}[q^*_{FFG}]$ &  & -104.42 & -86.61 &  -83.82 \\ \cline{1-1} \cline{3-5} 
% $\mathcal{L}_{\text{VAE}}[q_{FFG}]$   &  & TODO    & TODO   & TODO   \\ 
Approximation Gap                        &  & 3.90    & 1.83   & 1.63   \\
% Amortization                          &  & TODO    & TODO   & TODO   \\
% Inference Gap                         &  & TODO    & TODO   & TODO  
\end{tabular}
\caption{Increased decoder capacity reduces the approximation gap. }
\label{dec_table}
\end{table}

Given that we have seen that the generator can accommodate the choice of approximation, our next question is whether a generator with more capacity increases its ability to fit to the approximation.
To this end, we trained VAEs with decoders of different sizes and measured the approximation gaps on the training set. Specifically, we trained decoders with 0, 2, and 4 hidden layers on MNIST. See Table \ref{dec_table} for the results. We see that as the capacity of the decoder increases, the approximation gap decreases. This result implies that the more flexible the generator is, the less flexible the approximate distribution needs to be to ensure accurate inference.

\subsection{Inference Generalization}

How well does amortized inference generalize at test time? We address this question by visualizing the gaps
% Can we use the amortized IWAE bound on held-out data? 
% % Up till now we've considering how inference is affect during training on the training set. Now we will focus on inference on held-out data. 
% In Figure \ref{gaps_over_epochs}, we plot the gaps 
on training and validation datapoints across the training epochs. In Fig. \ref{gaps_over_epochs}, the models are trained on 50000 binarized Fashion-MNIST datapoints and the gaps are computed on a subset of a 100 training and validation datapoints. 
The top and bottom boundaries of the blue region represent $\log \hat{p}(x)$ and $\mathcal{L}[q^*]$. The bottom boundary of the orange region represents $\mathcal{L}[q]$.
% The top of the blue area is the estimate of $\log p(x)$, the bottom of the blue is $L[q^*]$, and the bottom of the orange is $L[q]$. 
In other words, the blue region is the approximation gap and the orange is the amortization gap.

% In Figure \ref{gaps_over_epochs}, we plot the gaps  on two subsets of data each consisting of 100 examples from the training and validation sets respectively. 
% The top and bottom boundaries of the blue region represent $\log \hat{p}(x)$ and $\mathcal{L}[q^*]$. The bottom boundary of the orange region represents $\mathcal{L}[q]$. In short, the blue and orange regions denote the approximation and amortization gaps respectively. 
In Fig. \ref{gaps_over_epochs}, the Standard model (top left) refers to a VAE of latent size 20 trained with a factorized Gaussian approximate posterior. In this case, the encoder and decoder both have two hidden layers each consisting of 200 hidden units. The Flow model (top right) augments the Standard model with a $q_{Flow}$ variational distribution. 
Larger Decoder and Larger Encoder models have factorized Gaussian distributions and increase the number of hidden layers to three and the number of units in each layer to 500.
% each hidden layer has number of units of the hidden layers to 500 and add a third hidden layer.

% respectively refer to models where the hidden layers of the decoder and encoder have 500 hidden units each. 
% All models are trained on a subset of 50000 examples and validated on a subset of 10000 examples from the binarized Fashion-MNIST dataset. 

Firstly, we observe that for all models, the approximation gap on the training and validation sets are roughly equivalent. This indicates that the true posteriors of the held-out data are similar to that of the training data. Secondly, we note that for all models, the encoder overfits more than the decoder. These observations resonate with the encoder overfitting findings by \citet{ais_eval}.
 
% On the other hand, the amortization gap on the validation set gradually becomes more prominent. This is due to the encoder overfitting to the training data, reducing its ability to generalize to the held-out data. 

How does increasing decoder capacity affect inference on held-out data? We know from Section \ref{approx_on_true} that increasing generator capacity results in a posterior that better fits the approximation making posterior inference easier. 
Furthermore, the Larger Decoder plot of Fig.  \ref{gaps_over_epochs} shows that increasing generator capacity causes the model to be more prone to overfitting. Thus, there is a tradeoff between ease of inference and decoder overfitting.

\subsubsection{Encoder Capacity and Approximation Expressiveness}
%point of this is that they both achieve similar logpx but flows's encoder is better because it can be used on teh test set , whereas the large encoder cant since its so overfit. 

We have seen in Sections \ref{amort_approx} and \ref{flows_on_amort} that expressive approximations as well as increasing encoder capacity can lead to a reduction in the amortization gap. This leads us to the following question: when should we increase encoder capacity versus increasing the expressiveness of the approximation? We answer this question in terms of how well each model can generalize its efficient inference (recognition network and variational distribution) to held-out data.  

In Fig. \ref{gaps_over_epochs}, we see that the Flow model and the Larger Encoder model achieve similar $\log \hat{p}(x)$ on the validation set at the end of training. However, we see that the $\mathcal{L}[q]$ bound of the Larger Encoder model is significantly lower than the $\mathcal{L}[q]$ bound of the Flow model due to the encoder overfitting to the training data. Although they both model the data nearly equally well, the recognition net of the Larger Encoder model is no longer suitable to perform inference on the held-out data due to overfitting. Thus a potential rational for utilizing expressive approximations is that they improve generalization to held-out data in comparison to increasing the encoder capacity.

We highlight that, in many scenarios, efficient test time inference is not required and consequently, encoder overfitting is not an issue, since we can use non-efficient encoder-independent methods to estimate $\log p(x)$, such as AIS, IWAE with local optimization, or potentially retraining the encoder on the held-out data. In contrast, when efficient test time inference is required, encoder generalization is important and expressive approximations are likely advantageous.

\subsection{Annealing the Entropy} \label{warmup_q}

Typical warm-up \citep{warmup1,sonderby2016ladder} refers to annealing the $\text{KL}\left( q(z|x)||p(z) \right)$ term during training. This can also be interpreted as performing maximum likelihood estimation (MLE) early on during training. This optimization technique is known to help prevent the latent variable from degrading to the prior \citep{iwae, sonderby2016ladder}. 
% keep the latent units "alive".
We employ a similar annealing scheme during training by annealing the entropy of the approximate distribution:
$$\mathbb{E}_{z \sim q(z|x)} \left[ \log p(x,z) - \lambda \log q(z|x) \right],$$
where $\lambda$ is annealed from 0 to 1 over training.  This can be interpreted as \textit{maximum a posteriori} (MAP)  in the initial phase of training. 
% Due to its similarity, we will also refer to this technique as warm-up.

We find that warm-up techniques, such as annealing the entropy, are important for allowing the true posterior to be more complex. Table \ref{larger_encoder} (right) are results from a model trained without the entropy annealing schedule. Comparing these results to Table \ref{opt_table}, we observe that the difference between  $\mathcal{L}_{\text{VAE}}[q^*_{FFG}]$ and $\mathcal{L}_{\text{VAE}}[q^*_{AF}]$ is significantly smaller without entropy annealing. This indicates that the true posterior is more Gaussian when entropy annealing is not used.
% The difference between the Optimal Flow and Optimal FFG for the Flow-trained model is 1.5 nats. In contrast, when trained with warm-up in Table \ref{opt_table}, the difference is 12.5 nats. Thus, the difference between the Optimal Flow and Optimal FFG is significantly smaller without warm-up than with warm-up. 
This suggests that, in addition to preventing the latent variable from degrading to the prior, entropy annealing allows the true posterior to better utilize the flexibility of the expressive approximation.
%, resulting in a better trained model.

% This suggests that warm-up allows the approximation to fit the true posterior early during training, which seems to not fit the FFG constraints.

% Table \ref{no_warmup} shows that if trained with FFG, the difference between the optimal FFG and Flow is only 0.1 nat, whereas for the flow trained model, the difference is 1.5 nat. Thus the FFG cant fit it as well to the Flow model as it can to the FFG model. This suggests that the true posterior is more similar to FFG when trained with FFG, and its more complex when trained with a more expressive approximation.

% Warmup is very important. It allows the posterior to be less Gaussian. 
% . Here we show that it is important for allowing the true posterior be complex, instead of Gaussian.

% \section{Conclusion/Discussion/Future Work}
\section{Conclusion}

In this paper, we investigated how encoder capacity, approximation choice, decoder capacity, and model optimization influence inference suboptimality in terms of the approximation and amortization gaps.
% broke down the inference gap into two parts: the amortization and approximation gaps. We explored how these gaps were affected by different datasets, different posterior approximations, and different encoder capacities. 
We discovered that the amortization gap can be a leading source to inference suboptimality and that the generator can reduce the approximation gap by learning a true posterior that fits to the choice of approximation. We showed that the parameters used to increase the expressiveness of the approximation play a role in generalizing inference rather than simply improving the complexity of the approximation. We confirmed that increasing the capacity of the encoder reduces the amortization error. Additionally, we demonstrated that optimization techniques, such as entropy annealing, help the generative model to better utilize the flexibility of expressive variational distributions. Analyzing these gaps can be useful for guiding improvements in VAEs. Future work includes evaluating other types of expressive approximations, more complex likelihood functions, and datasets.

\bibliography{icml2018_conference}
\bibliographystyle{icml2018}

\newpage

\twocolumn[
\icmltitle{Supplementary}
]

Code for the experiments in this paper can be found at: https://github.com/chriscremer/Inference-Suboptimality and https://github.com/lxuechen/inference-suboptimality. 

\subsection{Model Architectures and Training Hyperparameters}

%  $D_X$ refers to the dimensionality of the input and $D_Z$ refers to the dimensionality of the latent space. For MNIST and Fashion MNIST,  $D_X=784.$ 

\subsubsection{2D Visualization}

% The VAE model of Fig. \ref{true_post} used a decoder $p(x|z)$ with architecture: $D_Z-100-D_X$. The encoder $q(z|x)$ was $D_X-100-2D_Z$, where the latent space has dimensionality of 2. 
% % one hidden layer in each of the encoder and decoder networks with 100 units and 
% We used tanh activation functions and a batch size of 50. The model was trained for 3000 epochs with a learning rate of 0.0001. 

The VAE model of Fig. \ref{true_post} uses a decoder $p(x|z)$ with architecture: $2-100-784$, and an encoder $q(z|x)$ with architecture:  $784-100-4$. We use tanh activations and a batch size of 50. The model is trained for 3000 epochs with a learning rate of $10^{-4}$ using the ADAM optimizer \citep{kingma2014adam}. 

\subsubsection{MNIST \& Fashion-MNIST} \label{exps_details}
Both MNIST and Fashion-MNIST consist of a training and test set with 60000 and 10000 datapoints respectively, where each datapoint is a 28x28 grey-scale image. We rescale the original images so that pixel values are within the range $[0,1]$. For MNIST, We use the statically binarized version described by \cite{larochelle2008classification}. We also binarize Fashion-MINST \textit{statically}. For both datasets, we adopt the Bernoulli likelihood for the generator.

The VAE models for MNIST and Fashion-MNIST experiments have the same architecture.
%  given in table \ref{tab:mnist-arch}. 
The encoder has two hidden layers with 200 units each. The activation function is chosen to be the exponential linear unit (ELU, \citet{clevert2015fast}), as we observe improved performance compared to tanh. The latent space has 50 dimensions. The generator is the reverse of the encoder. We follow the same learning rate schedule and train for the same amount of epochs as described by \cite{iwae}. All models are trained with the a batch-size of 100 with ADAM.

% \begin{table}[ht]
%     \centering
%     \begin{tabular}{ c | c}
%         \textbf{Inference Network}  &  \textbf{Generator} \\
%         \hline
%         Input $\in \mathbb{R}^{784}$         & Input $\in \mathbb{R}^{50}$ \\
%         FC. 200-ELU-FC. 200-ELU-FC. 50+50    & FC. 200-ELU-FC. 200-ELU-FC. 784-Sigmoid \\
%     \end{tabular}
%     \caption{Neural net architecture for MNIST/Fashion-MNIST experiments.}
%     \label{tab:mnist-arch}
% \end{table}

% \begin{table}[b]
%     \centering
%     \begin{tabular}{ c | c}
%         $\bm{q(v_0|z_0)}$  & $\bm{r(v_T|z_T)}$ \\
%         \hline
%         Input $\in \mathbb{R}^{50}$         & Input $\in \mathbb{R}^{50}$ \\
%         FC. 100-ELU-FC. 100-ELU-FC. 50+50   & FC. 100-ELU-FC. 100-ELU-FC. 50+50 \\
%     \end{tabular}
%     \vspace{5mm}

%     \begin{tabular}{ c | c}
%     \multicolumn{2}{c}{$\bm{q(v_{t+1}, z_{t+1} | v_t, z_t)}$}\\
%     \hline
%         $\sigma_1(\cdot), \sigma_2(\cdot) $  &  $ \mu_1(\cdot), \mu_2(\cdot) $ \\
%         \hline
%         Input $\in \mathbb{R}^{50}$         & Input $\in \mathbb{R}^{50}$ \\
%         FC. 100-ELU-FC. 100-ELU-FC. 50      & FC. 100-ELU-FC. 100-ELU-FC. 50 \\
%     \end{tabular}
%     \caption{Flow setting for MNIST/Fashion-MNIST experiments. $q(v_T, z_T | v_0, z_0)$ consists of two normalizing flows given in the second tabular. }
%     \label{tab:flow}
% \end{table}

In the large encoder setting, we change the number of hidden units for the inference network to be 500, instead of 200. The warm-up models are trained with a linear schedule over the first 400 epochs according to Section \ref{warmup_q}.

% The flow configuration is given in table \ref{tab:flow}.
The auxiliary variable of  requires a couple distributions: $q(v_0|z_0)$ and $r(v_T|z_T)$. These distributions are both factorized Gaussians which are parameterized by MLP's with two hidden layers, 100 units each, with ELU activations. 

The flow transformation $q(z_{t+1},v_{t+1}|z_t,v_t)$ involves functions $\sigma_1$, $\sigma_2$, $\mu_1$, and $\mu_2$ from Eqn. \ref{trans1} and \ref{trans2}. These also have two hidden layers with 100 units each and ELU units.

\subsubsection{3-BIT CIFAR} \label{3bitcifar}
CIFAR-10 consists of a training and test dataset with 50000 and 10000 datapoints respectively, where each datapoint is a $32\times32$ RGB image. We rescale individual pixel values to be in the range $[0,1]$. 
% We follow the \textit{discretized logistic} likelihood model adopted by \cite{iaf}, where each input channel has its own scale learned by an MLP. 
% To simplify the problem and avoid numerical instability, 
We then statically binarize the scaled pixel values
% original rescaled CIFAR-10 dataset. 
% This means we 
by setting individual pixel values of channels to 1 if the rescaled value is greater than 0.5 and 0 otherwise. In this manner, we can model the observation with a factorized Bernoulli likelihood. We call this binarized CIFAR-10 dataset as \textit{3-BIT CIFAR}, since 3 bits are required to encode each pixel, where 1 bit is needed for each of the channels. We acknowledge that such binarization scheme may reduce the complexity of the original problem, since originally 24 bits were required to encode a single pixel. Nevertheless, the 3-bit CIFAR dataset is still much more challenging compared MNIST and Fashion. This is because 784 bits are required to encode one MNIST/Fashion image, whereas for one 3-bit CIFAR image, 3072 bits are required. Most notably, we were able to validate our AIS estimates using BDMC with the simplified dataset. This, however, was not achievable in any reasonable amount of time with the original CIFAR-10 dataset.

For the latent variable, we use a 50-dimensional factorized Gaussian for $q(z|x)$. For all neural networks, ELU is chosen to be the activation function. The inference network consists of three 4 by 4 convolution layers with stride 2, batch-norm, and 64, 128, 256 channels respectively. Then a fully-connected layer outputs the 50-dimensional mean and log-variance of the latent variable. Similarly, the generator consists of a fully-connected layer outputting 256 by 2 by 2 tensors. Then three deconvolutional layers each with 4 by 4 filters, stride 2, batch-norm, and 128, 64, and 3 channels respectively. For the model with expressive inference, we use three normalizing flow steps, where the parametric functions in the flow and auxiliary variable distribution also take in a hidden layer of the encoder.

% !Mention those other connections of the deconv layer

% \begin{table}[ht]
%     \centering
%     \begin{tabular}{ c | c }
%         \textbf{Inference Network} & \textbf{Generator}\\
%         \hline
%         Input $32\times 32$ color image                        & Input $\in \mathbb{R}^{32}$ \\
%         $4\times4$ conv. 64 channels. stride 2. BN. ELU           & FC. $256\times2\times2$ ELU \\
%         %\scriptsize{FC. $256\times2\times2$ ELU;} \scriptsize{FC. 64-ELU-FC. 32-ELU-FC. 3}\\
%         $4\times4$ conv. 128 channels. stride 2. BN. ELU          & $4\times4$ deconv. 128 channels. stride 2. BN \\
%         $4\times4$ conv. 256 channels. stride 2. BN.ELU          & $4\times4$ deconv. 64 channels. stride 2. BN \\
%         FC. $32+32$. mean and log-variance    & $4\times4$ deconv. 3 channels. stride 2. Sigmoid \\
%     \end{tabular}
%     \caption{Network architecture for CIFAR-10 experiments. 
%     %For the generator, one of the MLPs immediately after the input layer of the generator outputs channel-wise scales for the discretized logistic likelihood model. 
%     BN stands for batch-normalization.}
%     \label{tab:arch}
% \end{table}

We use a learning rate of $10^{-3}$. Warm-up is applied with a linear schedule over the first 50 epochs. All models are trained with a batch-size of 100 with ADAM. Early-stopping is applied based on the performance computed with the IWAE bound (k=1000) on the held-out set of 5000 examples from the original training set.

% as opposed to only two in MNIST/Fashion-MNIST experiments. 

% To simplify the problem and avoid numerical instability, we statically binarize the original rescaled CIFAR-10 dataset. This means we set individual pixel values of channels to 1 if the rescaled value is greater than 0.5 and 0 otherwise. In this manner, we can model the observation with a factorized Bernoulli likelihood. We call this binarized CIFAR-10 dataset as \textit{3-bit CIFAR}, since 3 bits are required to encode each pixel, where 1 bit is needed for each of the channels.

% We acknowledge that such binarization scheme may reduce the complexity of the original problem, since originally 24 bits were required to encode a single pixel. Nevertheless, the 3-bit CIFAR dataset is still much more challenging compared MNIST and Fashion. This is because 784 bits are required to encode one MNIST/Fashion image, whereas for one 3-bit CIFAR image, 3072 bits are required. 
% Most notably, we were able to validate our AIS estimates using BDMC with the simplified dataset. This, however, was not achievable in any reasonable amount of time with the original CIFAR-10 dataset.

\subsection{Inference Generalization} 

These models are trained with batch size 50 and latent dimension size of 20. The rest of the hyperparameters are equivalent to Section \ref{exps_details}.

Architecture of $q_{Flow}$: The flow transformation involves functions $\sigma_1$, $\sigma_2$, $\mu_1$, and $\mu_2$ from Eqn. \ref{trans1} and \ref{trans2}. Each function is an MLP with a 50 unit hidden layer and ELU activations. We apply this flow transformation twice.

\begin{figure}[ht]
  \centering
          \includegraphics[width=.51 \textwidth, clip, trim=1.8cm .3cm 1.cm 0.cm]{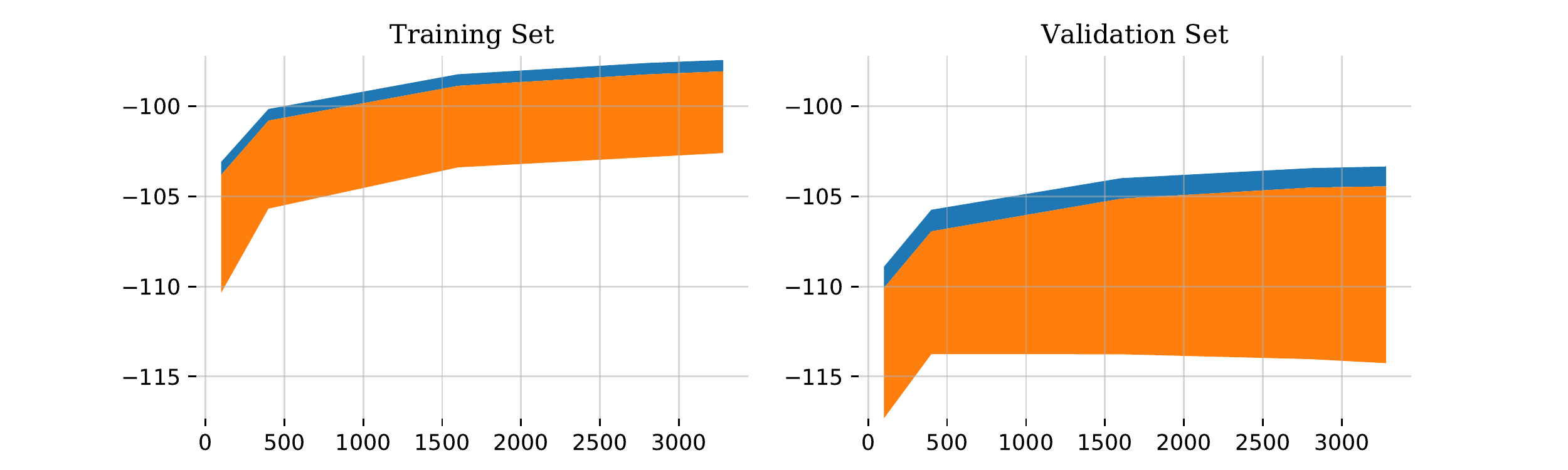} 
      \caption{Gaps over epochs of the AF (auxiliary flow) model.}
    %   \label{ais_iw}
    \label{af_gaps}
\end{figure}

Fig. \ref{af_gaps} are the plots for the $q_{AF}$ model. The transformations are the same as $q_{Flow}$, but rather than partitioning the latent variable, we introduce an auxiliary variable. The auxiliary variable also requires a reverse model $r(v|z)$ which is a factorized Gaussian parameterized by an MLP with a 50 unit hidden layer and ELU activations.

Comparing AF in Fig. \ref{af_gaps} to Flow in Fig. \ref{gaps_over_epochs}, we see that the AF has a larger approximation gap.
% , even though both models should be su
% In Fig. \ref{gaps_over_epochs}, we see that  $\mathcal{L}[q_{Flow}^*]$ is very close to $\log \hat{p}(x)$. 
% {\color{red}{TODO: elaborate, why is KL of reverse the issue. people might not recall the difference between flow and AF.}}
This increase is likely due to the 
% This indicates that the approximation gap of $q_{AF}$ could be dominated by the
% mainly due to the 
$\text{KL}\left(q(v|z,x)\|r(v|x,z)\right)$ term of the auxiliary variable lower bound from \ref{aux_section}. This motivates also using expressive approximations for the reverse model $r(v|z)$.

\subsection{Influence of Flows On Amortization Gap Experiment} \label{flow_encoder}

The aim of this experiment is to show that the parameters used for increasing the expressiveness of the approximation also contribute to reducing the amortization error. To show this, we train a VAE on MNIST, discard the encoder, then retrain two encoders on the fixed decoder: one with a factorized Gaussian distribution and the other with a parameterized 'flow' distribution. We use fixed decoder so that the true posterior is constant for both encoders. See \ref{flows_on_amort} for the results and below for the architecture details.

The architecture of the decoder is: $D_Z-200-200-D_X$.
The architecture of the encoder used to train the decoder is  $D_X-200-200-2D_Z$. The approximate distribution $q(z|x)$ is a factorized Gaussian.

Next, we describe the encoders which were trained on the fixed trained decoder. In order to highlight a large amortization gap, we employed a very small encoder architecture: $D_X-2D_Z$. This encoder has no hidden layers, which greatly impoverishes its ability and results in a large amortization gap. 

We compare two approximate distributions $q(z|x)$. Firstly, we experiment with the typical fully factorized Gaussian (FFG). The second is what we call a flow distribution. Specifically, we use the transformations of \cite{nvp}. We also include an auxiliary variable so we don't need to select how to divide the latent space for the transformations. The approximate distribution over the latent $z$ and auxiliary variable $v$ factorizes as: $q(z,v|x)=q(z|x)q(v)$. The $q(v)$ distribution is simply a N(0,1) distribution. Since we're using a auxiliary variable, we also require the $r(v|z)$ distribution which we parameterize as $r(v|z)$: $[D_Z]-50-50-2D_Z$. The flow transformation is the same as in Section \ref{aux_nf}, which we apply twice.

\subsection{Computation of the Determinant for Flow} \label{flow_det}
The overall mapping $f$ that performs $(z,v) \mapsto (z',v')$ is the composition of two sheer mappings $f_1$ and $f_2$ that respectively perform $(z,v) \mapsto (z,v')$ and $(z,v') \mapsto (z',v')$. Since the Jacobian of either one of the sheer mappings is diagonal, the determinant of the composed transformation's Jacobian $Df$ can be easily computed:
% $$ \mathrm{det} (Df) = \mathrm{det}(Df_1)  \mathrm{det}(Df_2) = \Bigl(\prod_{i=1}^n \sigma_1(z)_i \Bigr) \Bigl(\prod_{j=1}^n\sigma_2(v')_j \Bigr).$$
\begin{align*}
    \mathrm{det} (Df) &= \mathrm{det}(Df_1)  \mathrm{det}(Df_2) \\
    &= \Bigl(\prod_{i=1}^n \sigma_1(z)_i \Bigr) \Bigl(\prod_{j=1}^n\sigma_2(v')_j \Bigr).
\end{align*}

\subsection{Annealed Importance Sampling} \label{ais_section}
Annealed importance sampling (AIS, \citet{ais, jarzynski1997nonequilibrium}) is a means of computing a lower bound to the marginal log-likelihood. Similarly to the importance weighted bound, AIS must sample a proposal distribution $f_1(z)$ and compute the density of these samples, however, AIS then transforms the samples through a sequence of reversible transitions $\mathcal{T}_t(z'|z)$. The transitions anneal the proposal distribution to the desired distribution $f_T(z)$. 

Specifically, AIS samples an initial state $z_1 \sim f_1(z)$ and sets an initial weight $w_1 =1$. For the following annealing steps, $z_t$ is sampled from $\mathcal{T}_t(z'|z)$ and the weight is updated according to:
$$w_t = w_{t-1} \frac{f_t(z_{t-1})}{f_{t-1}(z_{t-1})}.$$ 

This procedure produces weight $w_T$ such that $\mathbb{E} \left[w_T \right] =  \mathcal{Z}_T / \mathcal{Z}_1$, where $Z_T$ and $Z_1$ are the normalizing constants of $f_T(z)$ and $f_1(z)$ respectively. This pertains to estimating the marginal likelihood when the target distribution is $p(x,z)$ when we integrate with respect to $z$.

Typically, the intermediate distributions are simply defined to be geometric averages: $f_t(z) = f_1(z)^{1-\beta_t} f_T(z)^{\beta_t}$, where $\beta_t$ is monotonically increasing with $\beta_1$ = 0 and $\beta_T$ = 1. When $f_1(z) = p(z)$ and $f_T(z) = p(x,z)$, the intermediate distributions are:
$f_i(x) = p(z) p(x|z)^{\beta_i}.$ 

Model evaluation with AIS appears early on in the setting of deep belief networks \citep{salakhutdinov2008quantitative}. AIS for decoder-based models was also used by \citet{ais_eval}.
% They validated the accuracy of the approach with Bidirectional Monte Carlo (BDMC, \citet{grosse2015sandwiching}) and demonstrated the advantage of using AIS over the IWAE bound for evaluation when the inference network overfits to the training data.

\subsection{Extra MNIST Inference Gaps}

\begin{figure}[ht]
  \centering
    %   \subfloat[1000 datapoints]
    %     {
        \includegraphics[width=.5 \textwidth]{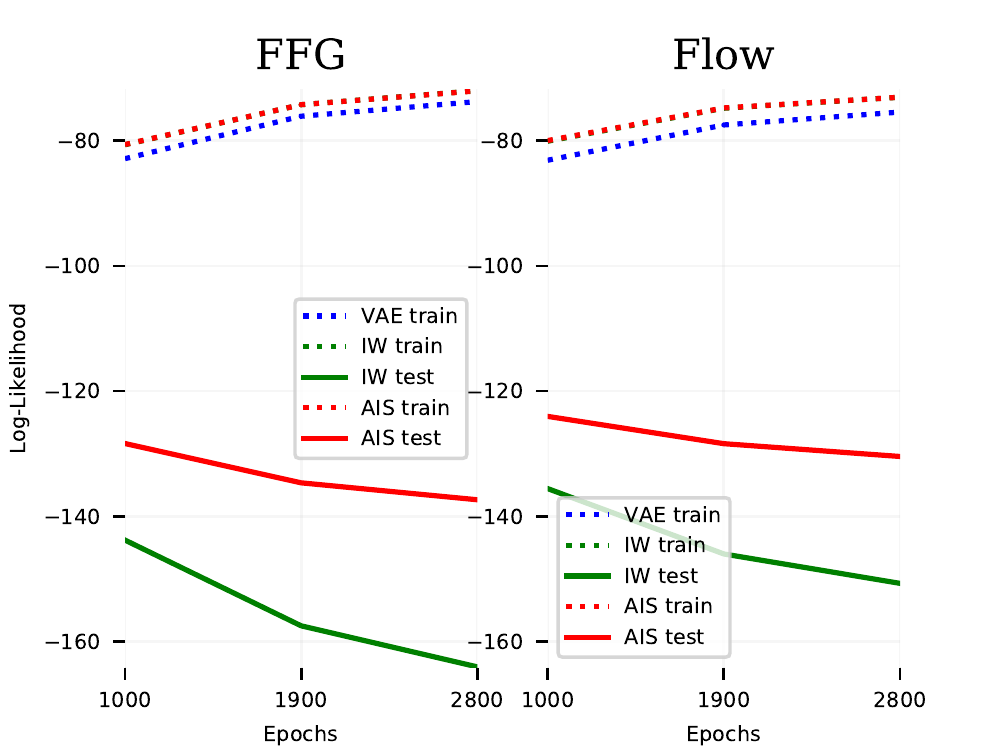}
        % }
      \caption{Training curves for a FFG and a Flow inference model on MNIST. AIS provides the tightest lower bound and is independent of encoder overfitting. There is little difference between FFG and Flow models trained on the 1000 datapoints since inference is nearly equivalent.}
       \label{smallN}
    %   \label{ais_iw}
\end{figure}

To demonstrate that a very small inference gap can be achieved, even with a limited approximation such as a factorized Gaussian, we train the model on a small dataset. In this experiment, our training set consists of 1000 datapoints randomly chosen from the original MNIST training set. The training curves on this small datatset are shown in Fig. \ref{smallN}. Even with a factorized Gaussian distribution, the inference gap is very small: the AIS and IWAE bounds are overlapping and the VAE is just slightly below. Yet, the model is overfitting as seen by the decreasing test set bounds. % This leads to the next topic of encoder and decoder overfitting.

% Test set inference
    % encoder overfitting 
    % decoder overfitting 

% \begin{figure}[ht]
%   \centering
%       \subfloat[Full dataset]
%         {
%         %   \includegraphics[width=.5 \textwidth]{figs/ais_iw_just_vae_flow.png}
%           \includegraphics[width=.5 \textwidth]{figs/plot_126.pdf} \label{largeN}
%         }
%       \subfloat[1000 datapoints]
%         {
%         \includegraphics[width=.5 \textwidth]{figs/plot_125.pdf} \label{smallN}
%         }
%       \caption{Training curves for a FFG and a Flow inference model on MNIST. AIS provides the tightest lower bound and is independent of encoder overfitting. There is little difference between FFG and Flow models trained on the 1000 datapoints since inference is nearly equivalent.}
%       \label{ais_iw}
% \end{figure}

% \begin{figure}[ht]
%   \centering
%     %   \subfloat[Full dataset]
%     %     {
%         %   \includegraphics[width=.5 \textwidth]{figs/ais_iw_just_vae_flow.png}
%           \includegraphics[width=.5 \textwidth]{figs/plot_126.pdf} 
%       \caption{Training curves for a FFG and a Flow inference model on MNIST. AIS provides the tightest lower bound and is independent of encoder overfitting. There is little difference between FFG and Flow models trained on the 1000 datapoints since inference is nearly equivalent.}
%     %   \label{ais_iw}
%     \label{largeN}
% \end{figure}

\end{document}